\documentclass[runningheads]{llncs}
\usepackage[T1]{fontenc}
\usepackage{graphicx}
\usepackage{hyperref}
\usepackage{color}

\usepackage{comment}
\usepackage{amsmath}
\DeclareMathOperator*{\argmin}{arg\,min}
\usepackage{amssymb}
\begin{document}
\title{Spatio-temporal neural distance fields for conditional generative modeling of the heart}
\titlerunning{Spatio-temporal generative modeling}
% If the paper title is too long for the running head, you can set
% an abbreviated paper title here
%
\author{Kristine S\o rensen\inst{1}\and
Paula Diez\inst{1} \and
Jan Margeta\inst{2} \and
Yasmin El Youssef\inst{3} \and
Michael Pham\inst{3} \and
Jonas Jalili Pedersen\inst{3} \and
Tobias K{\"u}hl\inst{3,4} \and
Ole de Backer\inst{3} \and
Klaus Kofoed\inst{3} \and
Oscar Camara\inst{5} \and
Rasmus Paulsen\inst{1}}
% index{Sørensen, Kristine}
% index{Diez, Paula}
% index{Margeta, Jan}
% index{El Youssef, Yasmin}
% index{Pham, Michael}
% index{Pedersen, Jonas Jalili}
% index{Kühl, Tobias}
% index{de Backer, Ole}
% index{Kofoed, Klaus}
% index{Camara, Oscar}
% index{Paulsen, Rasmus}

%
\authorrunning{K. S\o rensen et al.}
% First names are abbreviated in the running head.
% If there are more than two authors, 'et al.' is used.
%
\institute{DTU Compute, Technical University of Denmark, Kongens Lyngby, Denmark \and 
KardioMe, Research \& Development, Nova Dubnica, Slovakia \and
Heart center, Rigshospitalet, Copenhagen, Denmark \and
Dep. of Cardiology, Zealand University Hospital, Denmark \and
Physense, BCN MedTech, Universitat Pompeu Fabra, Barcelona, Spain\\
\email{kajul@dtu.dk}}
\maketitle              % typeset the header of the contribution
\begin{abstract}
    The rhythmic pumping motion of the heart stands as a cornerstone in life, as it circulates blood to the entire human body through a series of carefully timed contractions of the individual chambers. 
Changes in the size, shape and movement of the chambers can be important markers for cardiac disease and modeling this in relation to clinical demography or disease is therefore of interest. 
Existing methods for spatio-temporal modeling of the human heart require shape correspondence over time or suffer from large memory requirements, making it difficult to use for complex anatomies. 
We introduce a novel conditional generative model, where the shape and movement is modeled implicitly in the form of a spatio-temporal neural distance field and conditioned on clinical demography. 
The model is based on an auto-decoder architecture and aims to disentangle the individual variations from that related to the clinical demography. 
It is tested on the left atrium (including the left atrial appendage), where it outperforms current state-of-the-art methods for anatomical sequence completion and generates synthetic sequences that realistically mimics the shape and motion of the real left atrium. 
In practice, this means we can infer functional measurements from a static image, generate synthetic populations with specified demography or disease and investigate how non-imaging clinical data effect the shape and motion of cardiac anatomies. 

\end{abstract}
\section{Introduction}
    During a heart beat, the heart chambers undergo a series of complex 3D deformations and the morphology as well as the motion of contraction and relaxation are critical functions to pump blood to all human organs. 
Such shape and motion can be captured by Computed Tomography (CT), where Cardiac Functional Analysis (CFA) has enabled fast acquisition with 20 (or more) time frames per heartbeat.  
The cardiac anatomy and motion vary widely between individuals and it is therefore of interest to disentangle the individual characteristics from that of clinical factors such as gender, age or disease. 
% This is included in the final version
Traditional shape modeling based on shape correspondence and point distribution models \cite{cootes1995active} can be used for generative modeling, but the integration of clinical data in such models is currently an unsolved problem. 
Modeling the spatio-temporal characteristics of the heart has been achieved through 4D registration methods \cite{peyrat2010} or by building spatio-temporal atlases \cite{hoogendoorn2009,medrano-gracia2014}. 
% Include a bit more shape modeling litterature here Cootes95, Slipsager2018 - cannot handle integration of demography data
More recently, deep learning based methods have been used to learn the distribution over plausible cardiac shapes based on dense point clouds \cite{peng2024,beetz2022c}, meshes \cite{beetz2022a} and voxel volumes \cite{biffi2020}. 
Modeling the temporal movement is usually approached by mapping the anatomy at one time frame to another time frame (ie. end-diastole (ED) to end-systole (ES) or vice versa) and has been handled using i.e. mesh U-nets \cite{beetz2022b}, spatio-temporal graph convolutions \cite{lu2021b} or adversarial methods in the image domain \cite{ossenberg-engels2020}. 
%Integrating non-imaging clinical data into the modeling has proven difficult for the more traditional methods, but has made its way into image generation frameworks 
Generative frameworks have incorporated non-imaging clinical data with methods for generating cardiac ultrasound images with specified functional properties \cite{reynaud2022} and Magnetic Resonance Images (MRI) with specified anatomical characteristics \cite{amirrajab2023}, pathology \cite{duchateau2018} or biophysical parameters \cite{prakosa2013}. 
Generation of spatio-temporal cardiac anatomies based on clinical demography has been approached by \cite{qiao2023}, who learned a temporal latent space based on a recurrent neural network and generated cardiac anatomies represented by voxelized labelmaps. 
The explicit voxelmap representation however suffer from large memory requirements and are not ideal for representing the smooth and often highly detailed cardiac anatomy. 
Implicit representations (such as distance fields) have shown to be an effective representation of complex shapes for medical image segmentation \cite{sorensen2022,amiranashvili2022,stolt-anso2023} as well as shape generation in 3D \cite{park2019,chou2023} and 4D \cite{erkoc2023}. 
In contrast to other representations, the continuous nature of neural distance fields allows for modeling highly complex structures without requiring correspondence between samples or have memory requirements growing cubically with the image resolution. 

We propose a novel conditional generative model architecture, that integrates clinical demographic data into the generation of spatio-temporal signed distance fields (SDFs) of dynamic anatomies. 
The model is tested for modeling the left atrium (LA) and its complex extrusion; the left atrial appendage (LAA). 
The shape and motion of the LA and LAA are related to the risk of thrombus formation and stroke \cite{Glikson2020} and a detailed representation of the spatio-temporal dynamics are therefore of interest. 
Capturing the small details of the LAA anatomy would require a very large voxel map and obtaining point correspondence across the widely varying samples are unattainable \cite{slipsager2018}. 
The use of a neural SDF representation allows for modeling complex surfaces jointly in space and time, while the auto-decoder formulation enables the derivation of two separate latent spaces that disentangles the patterns related to the clinical demography from those related to individual variation. 
The proposed method are used to complete the full cardiac cycle based on a single time frame and to generate realistic synthetic anatomical sequences with specified clinical demography. %\footnote{Code available at \url{https://github.com/kristineaajuhl/spatio_temporal_generative_cardiac_model.git}}

\section{Materials and methods}
    \noindent \textbf{Overall framework}\footnote{Code available at \url{https://github.com/kristineaajuhl/spatio_temporal_generative_cardiac_model.git}}\\
A spatio-temporal neural distance field is a neural function that maps an arbitrary space-time coordinate $\textbf{x} = (\textbf{p},t)$, consisting of the 3D coordinate $\textbf{p}$ and the time index $t$, to the signed distance $\hat{d} = f_\theta(\textbf{x})$. 
The surface represented by the neural SDF can thus be considered the decision boundary separating SDF$<$0 from SDF$>$0. 
To learn such decision boundary for a single sequence, the network is presented with a set $X$ consisting of space-time coordinates $\textbf{x}$ and corresponding SDF-values $d$, such that $ X := \{(\textbf{x},d) : \text{SDF}(\textbf{x}) = d \}$.
The network parameters $\theta$ are optimized using the clamped L1-loss between the true and predicted SDF-values in this set: 

\begin{equation}\label{eq:L1}
        \mathcal{L}(f_\theta(\textbf{x}),d) = |\textrm{clamp} (f_\theta(\textbf{x}), \delta) - \textrm{clamp}(d, \delta)|
\end{equation}

\noindent where the clamping parameter $\delta$ controls the distance from the surface, over which we expect to learn an accurate distance field. 
The continuous spatio-temporal SDF can then be approximated as $SDF(\textbf{S}) \approx f_\theta( \textbf{S})$, where $\textbf{S}$ refers to all possible space-time coordinates.

\begin{figure}[tb]
    \centering
    \includegraphics[width=\linewidth]{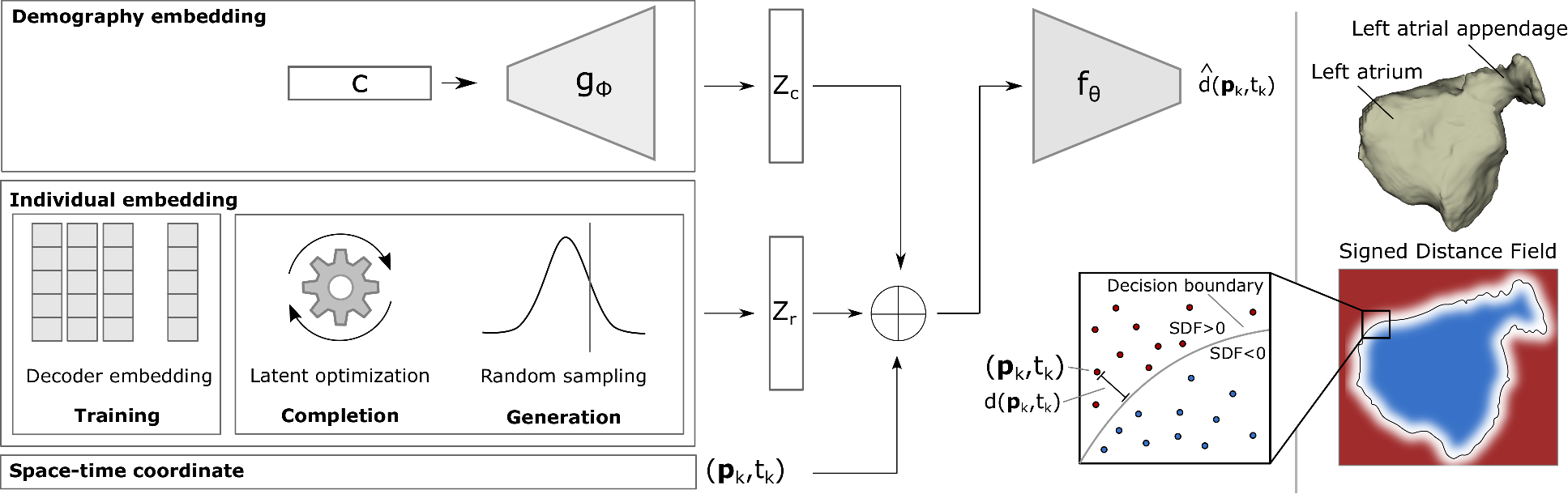}
    \caption{The signed distance field approximates the surface as the decision boundary separating space-time coordinates ($\textbf{p}_k,t_k$) that are inside and outside the surface. For each $\textbf{p}_k,t_k$ the signed distance $\hat{d}$ to the surface is predicted with the network $f_\theta$ based on a concatenation ($\oplus$) of the clinical demography latent vector $\textbf{z}_c$, the individual latent vector $\textbf{z}_r$ and the coordinate. $\textbf{z}_c$ is embedded from the clinical demography encoder $g_\phi$, whereas the source of $\textbf{z}_r$ depends on the task. \textbf{Training:} Each training sample is assigned a learnable embedding $\textbf{z}_r$ which is optimized jointly with $g_\phi$ and $f_\theta$. \textbf{Reconstruction:} The individual embedding is learned by locking the parameters of $f_\theta$ and $g_\phi$ and optimize for $\textbf{z}_r$. \textbf{Generation:} A new $\textbf{z}_r$ is generated by sampling from a multivariate Gaussian distribution. }
    \label{fig:pipeline}
\end{figure}

To model multiple anatomical sequences with the same neural network, we introduce a latent vector to represent each of the anatomical sequences indexed by $n=\{1,...,N\}$. 
This latent vector is concatenated with the space-time coordinate and given as input to $f_\theta$, where it enables the neural function to create decision boundaries unique to each anatomical sequence. 
We propose to learn this latent vector as two separate parts: the clinical demographic latent vector $\textbf{z}_c$ representing an embedding of the clinical demography (gender, age and SBP) and a residual latent vector $\textbf{z}_r$ representing the individual information that cannot be described by the clinical demography. 
$\textbf{z}_c$ is embedded from the clinical demography $\textbf{c}$ with a neural network such that $\textbf{z}_{c} = g_\phi(\textbf{c})$. 
The residual latent vector $\textbf{z}_{r}$ is a learnable embedding unique to each of the sequences in the training set ($\textbf{z}_{r,1},\textbf{z}_{r,2},...,\textbf{z}_{r,N})$ and is optimized jointly with $f_\theta$ and $g_\phi$ using an auto-decoder formulation. 
% THIS IS ADDED IN THE FINAL PAPER
The use of an auto-decoder circumvents the need for designing and training a 3D data encoder, but requires test time optimization which is slightly more time consuming and may risk converging to a local solution. 
To optimize the autodecoder parameters, we sample $K=110.000$ space-time coordinates for the 20 time frames ($t=\{0\%,5\%,...,95\%\}$ of the cardiac cycle) in each of the $N=290$ anatomical sequences in the training set and denote each of these samples $\textbf{s}_{n,k,t}$ and the corresponding SDF value $d_{n,k,t}$. 
The loss over the predicted and measured SDF-values across all N $\cdot$ T $\cdot$ K samples are computed as:

\begin{equation}
    \argmin_{\theta, \phi, \{\textbf{z}_{r,n}\}^{N}_{n=1}}\ \sum^{N}_{n=1} \sum^{K}_{k=1} \sum^{95\%}_{t=0\%} \mathcal{L}(f_\theta( g_\phi(\textbf{c}_n) \oplus \textbf{z}_{r,n} \oplus \textbf{s}_{n,k,t}),d_{n,k,t}) + \frac{1}{\sigma^2} || \textbf{z}_{r,n}||_2^2, 
\end{equation}

\noindent where the loss $\mathcal{L}$ is the clamped L1-loss from Equation \ref{eq:L1}, whereas $|| \textbf{z}_{r,n}||_2^2$ is a regularization term that encourages compact latent spaces, and $\sigma^2$ balances the L1-loss and the regularization. 
Figure \ref{fig:pipeline} shows an overview of how we propose to integrate neural SDFs in a conditional generative framework and visualizes an example of how the left atrial shape is encoded as the decision boundary of the neural network based on point-distance samples. 

%\textbf{Sequence completion $\cdot$}
\textbf{Sequence completion  ---}
The proposed model can complete an anatomical sequence based on the clinical demography $\textbf{c}$ and the static anatomy at a given time frame $t_{\text{given}}$. 
The clinical demography are mapped to the clinical demographic latent space as $\textbf{z}_{c} = g_\phi(\textbf{c})$, whereas the individual latent vector $\textbf{z}_r$ is found using test-time optimization. 
We sample $K=110.000$ space coordinates ($\textbf{s}_{k|t_\text{given}}$), measure the distances to the static anatomy at $t_\text{given}$ and optimize over the values in $\textbf{z}_r$ while locking the parameters of $f_\theta$ and $g_\phi$:

\begin{equation}
    \hat{\textbf{z}}_r = \argmin_{\textbf{z}_r}\ \sum^K_{k=1} \mathcal{L}(f_\theta( g_\phi(\textbf{c}) \oplus \textbf{z}_{r} \oplus \textbf{s}_{k|t_\text{given}}),d_{k|t_\text{given}}) + \frac{1}{\sigma^2} || \textbf{z}_{r}||_2^2, 
\end{equation}
 
The full anatomical sequence can be reconstructed by evaluating the SDF at all points on a uniform grid for the desired time frames and extract the zero-level isosurface using Marching Cubes \cite{lorensen1987marching}. 

%\subsection*{Sequence generation}
\textbf{Sequence generation ---}
The model can generate new plausible anatomical sequences by sampling randomly in the individual latent space $\textbf{z}_r$ and concatenate it with the embedding of the clinical demography. 
In the derivation of the auto-decoder formulation of neural distance fields (see \cite{park2019}) the prior distribution over the latent variables $\textbf{z}_r$ is assumed to come from a zero-mean multivariate Gaussian distribution. 
To generate new sequences with individual shape and motion characteristics, we sample $\textbf{z}_r$ from such distribution.
%We therefore fit such distribution to the individual latent vectors $\textbf{z}_r$ from the training set, and sample new unique individual latent vectors hereof. 

\noindent \textbf{Implementation details}\\
The model was implemented with PyTorch \cite{paszke2019} and trained on a NVIDIA RTX A4000 (16GB) for 9 hours.
The clinical demographic input is passed as a vector with one-hot encoded gender (male/female) and age-group ($<$50,50-59,60-69 and $>$69) and uses the continuous SBP normalized to $[0;1]$. 
Both $f_\theta$ and $g_\phi$ are implemented as a multi-layer perceptron (MLP). 
$f_\theta$ follows the architecture from \cite{park2019}, whereas $g_\phi$ consist of two hidden layers with each 128 neurons. 
We use 64 dimensions for both $\textbf{z}_c$ and $\textbf{z}_r$.
During training we randomly dropout $\textbf{z}_r$ in $20\%$ of the training steps, which was found necessary to force the network to make use of the clinical demography.
All surfaces are extracted from an SDF evaluated on an uniform grid sized $128^3$.

\noindent \textbf{Data and preprocessing}\\
A dataset of 4D geometries was extracted from CFA scans of 667 randomly selected participants from the Copenhagen General Population Study. 
All subjects are asymptomatic individuals from the general population, with 301/366 male/female participants, aged 41-89 years and with known systolic blood pressure (SBP).
Participation was conducted following the declaration of Helsinki and approved by the ethical committee (H-KF-01-144/01).
Each CFA series consists of a cardiac computed tomography angiography (CCTA) image at 20 equally spaced time steps during one heart beat ($t=0\%,5\%,...,95\%$). 
Each image was segmented using an automatic deep learning based segmentation method specifically developed for LAA segmentation from CT images \cite{sorensen2022}. 
We aligned all anatomies at $t=0\%$ by matching the center-of-mass, fine-tuned with a rigid transformation based on iterative closest point (ICP)\cite{arun1987} and applied the found transformation to all time frames. 
Finally, each anatomy was scaled with a fixed factor such that all surfaces lie within the unit-sphere.
We follow the sampling strategy from \cite{sorensen2022} and sample 10.000 random samples within the unit sphere and 100.000 samples in the vicinity of the surface based on the shape diameter at each vertex for each time frame. 
%This totals to $K=110k$ samples per person per time instance. 
All transformations and point-to-surface distances were obtained using VTK \cite{VTK}.
We split the dataset randomly with 290/10/367 anatomical sequences for training/validation/testing. 
%All subjects are asymptomatic individuals from the general population, with 301/366 male/female participants, aged 41-89 years and with a systolic blood pressure in the interval 84-192 mmHg. 
\section{Experiments and results}
    We evaluate the proposed methods on two different tasks - \textit{anatomical sequence completion} based on a single static anatomy as well as the clinical demography and \textit{anatomical sequence generation} based only on the clinical demography. 
%For sequence completion the whole cardiac cycle is generated based on a single static anatomy and the clinical demography of the person, whereas anatomical sequence generation creates unique sequences of unseen individuals based only on clinical demography. 

\textbf{Sequence completion ---}
Evaluating the sequence completion abilities of a generative model assure us that the model has learned a descriptive distribution of both shape and movement from the sequences in the training data. 
We report results for completion based on the static anatomy at $t=0\%$ as this allows for comparison to \cite{qiao2023}.
We evaluate the completion quality using symmetric Chamfer Distance (CD) and Hausdorff Distance (HD) between the predicted and true surfaces at each time step. 
The ability to capture the dynamic movement is measured by comparing the maximum volume ($V_\text{max}$), Fractional Change ($FC = (V_\text{max}-V_\text{min})/V_\text{max})$) and Cyclic Change $CC=V_\text{max}-V_\text{min}$ between the true and completed sequences.
%The reconstruction error is not constant during the heart beat, and shows to be lowest at the time frame from which we are completing the sequence (in this case $t=0$) and higher in time frames where the deformation to $t=0$ is large. 
%To reflect this difference we report the average CD and HD across all time frames as well as the minimum and maximum distances as an average across the test set. 
We have compared against the default CHeart implementation \cite{qiao2023} trained on our dataset as well as investigated the effect of removing the clinical demography from our model, by setting the clinical demography vector $\textbf{c}$ to a zero-vector for all samples during training and testing.
The results are seen in Table \ref{tab:reconstruction}.
We observe that the proposed method significantly outperforms CHeart \cite{qiao2023}, which is attributed the joint spatial and temporal modeling and the increased representational power from the SDF. 
An ablation study on the clinical demography, demonstrate a positive effect on estimating the functional parameters (FC, CC and $V_{max}$) even though the reconstruction errors do not show an improvement.

\begin{table}[b!]
    \centering
    \caption{Sequence completion evaluation measuring the Chamfer distance (CD), Hausdorff distance (HD), Maximum volume ($V_\text{max}$), Fractional Change (FC) and Cyclic Change (CC). For each test sample we collect the time instance with the minimum, average and maximum CD and HD and report the average across all test samples for each of these. Bold indicates best evaluation.}
    \label{tab:reconstruction}
    \begin{tabular}{|l|ccc|ccc|ccc|} \hline
                             Method & \multicolumn{3}{|c|}{CD [mm]} & \multicolumn{3}{|c|}{HD [mm]} &  \multicolumn{3}{|c|}{Difference [$\%$]} \\
                             & Min & Mean & Max & Min & Mean & Max & $V_\text{max}$  & FC & CC\\ \hline
    CHeart \cite{qiao2023}    & 2.80 & 4.13 & 6.23 & 9.59 & 16.79 & 54.16 & 4.41$\%$ & 8.26$\%$ & 9.65$\%$\\
    Ours ($\div$ demography)     & \textbf{1.71} & \textbf{2.88} & 4.26 & \textbf{6.75} & 11.64 & 17.08 & 3.90$\%$ & 5.64$\%$ & 9.28$\%$\\
    Ours                      & 1.93 & 2.89 & \textbf{4.08} & 6.94 & \textbf{11.49} & \textbf{16.56} & \textbf{3.32$\%$} & \textbf{4.59$\%$} & \textbf{7.83$\%$}  \\
    \hline 
    \end{tabular}
\end{table}

%\begin{table}
%    \centering
%    \caption{Results for evaluating the methods abilities for completing a sequence from a single time frame. Chamfer distance (CD), Hausdorff distance (HD), Maximum volume ($V_\text{max}$), Fractional Change ($FC = (V_\text{max}-V_\text{min})/V_\text{max})$) and Cyclic Change $CC=V_\text{max}-V_\text{min}$. For the distances we report both minimum, average and maximum distances with respect to the time instance and average across all test samples. Bold indicates best evaluation.}
%    \label{tab:reconstruction}
%    \begin{tabular}{|l|cc|ccc|} \hline
%                            & CD [mm] & HD[mm] & $V_\text{max}$ & FC & CC\\ \hline
%    Our                     & 2.89 (1.93-\textbf{4.08}) & \textbf{11.49} (6.94-\textbf{16.56}) & \textbf{3.32$\%$} & \textbf{4.59$\%$} & \textbf{7.83$\%$} \\
%    Our (unconditional)     & \textbf{2.88} (\textbf{1.71}-4.26) & 11.64 (\textbf{6.75}-17.08) & 3.90$\%$ & 5.64$\%$ & 9.28$\%$\\
%    CHeart \cite{qiao2023}  & 4.13 (2.80-6.23) & 16.79 (9.59-54.16) & 4.41$\%$ & 8.26$\%$ & 9.65$\%$ \\ \hline 
%    \end{tabular}
%\end{table}

\begin{figure}[t!]
    \centering
    \includegraphics[width=\linewidth]{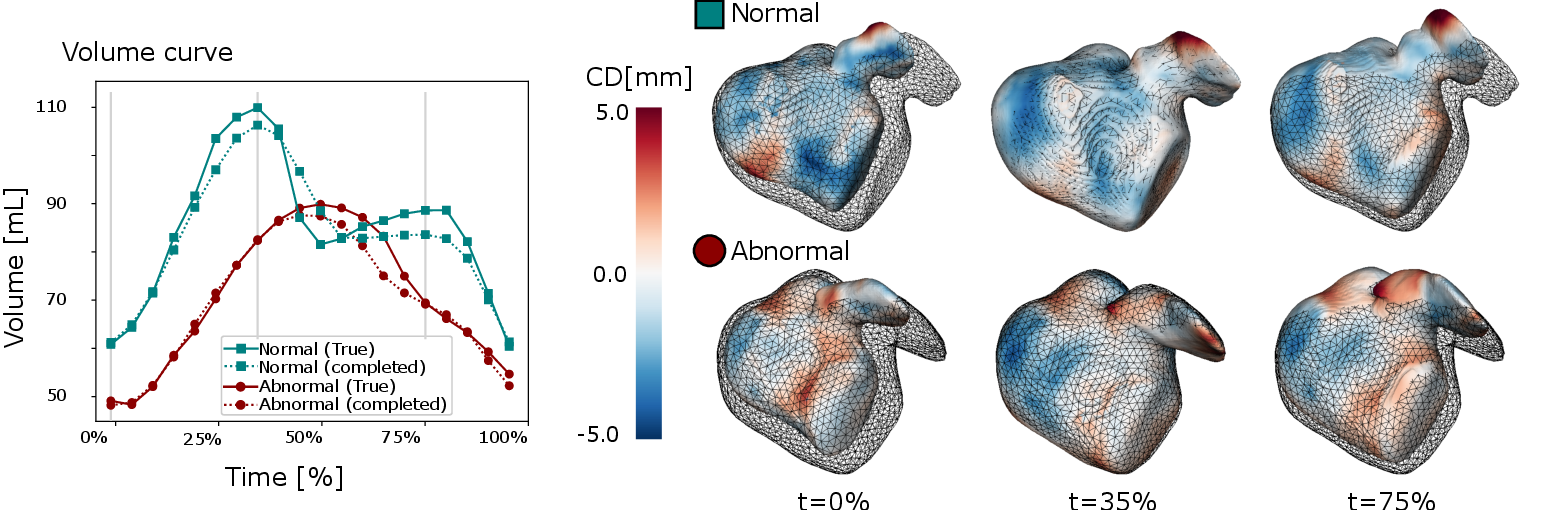}
    \caption{Two examples of completed sequences illustrated with volume curves and the chamfer distance (CD) between the predicted and true surface at time frame $0\%$,$35\%$ and $75\%$ (coloured surfaces) as well as surface with maximum volume (wireframe). The blue ($\square$) correspond to the $25th$ percentile evaluated on average CD, whereas the red ($\bigcirc$) shows an abnormal atrial motion without an active emptying phase.}
    \label{fig:reconstruction}
\end{figure}

Figure \ref{fig:reconstruction} shows the completion results from three different test cases. 
The blue volume curve is an example of a normal atrial function consisting of a relaxation phase ($t=0\%$ to first peak), a passive emptying phase (first peak to plateau at $t\approx 75\%$) and an active emptying phase (plateau to $t=95\%$). 
The red curve is an example of an abnormal atrial motion dominated by passive emptying. 
We observe that the proposed model correctly completes both of these sequences, which indicates that the model are able to learn individual motion patterns. 
%Since this timing is closely related to the heart rate and the heart rate is measured during scanning, we expect that this can be improved by adding the heart rate to the conditions.
%The well-completed abnormal atrial motion further indicates that the model are capable of learning different motion patterns dependent on the anatomy at $t=0$ and/or the clinical conditions. 

\begin{figure}[b!]
    \centering
    \includegraphics[width=\linewidth]{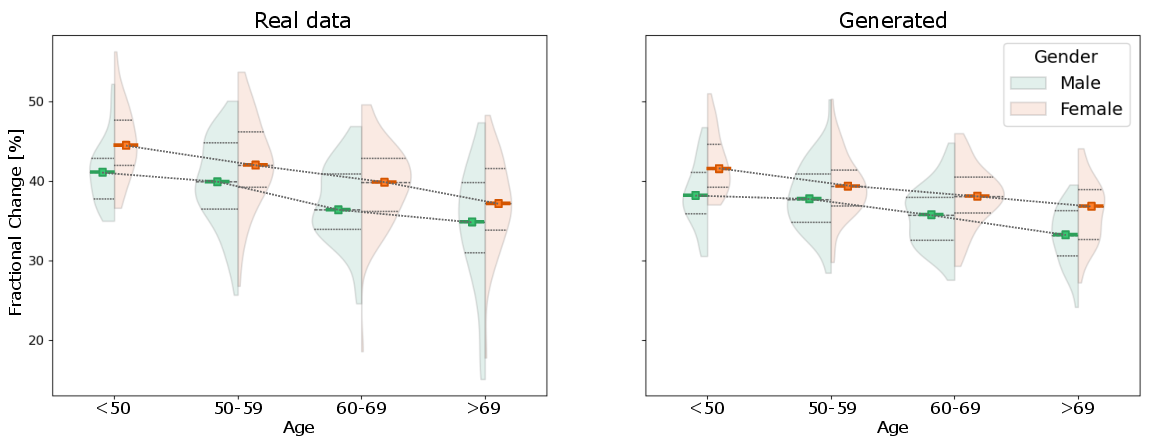}
    \caption{Distribution of left atrial fractional change (FC) across the subgroups based on gender (male/female) and age ($<50$,50-59,60-69 and $>69$) in the population from the test set (left) and a synthetic population generated with the same clinical demography as the population in the test set (right).}
    \label{fig:generated_distributions}
\end{figure}

\begin{figure}[b!]
    \centering
    \includegraphics[width=\linewidth]{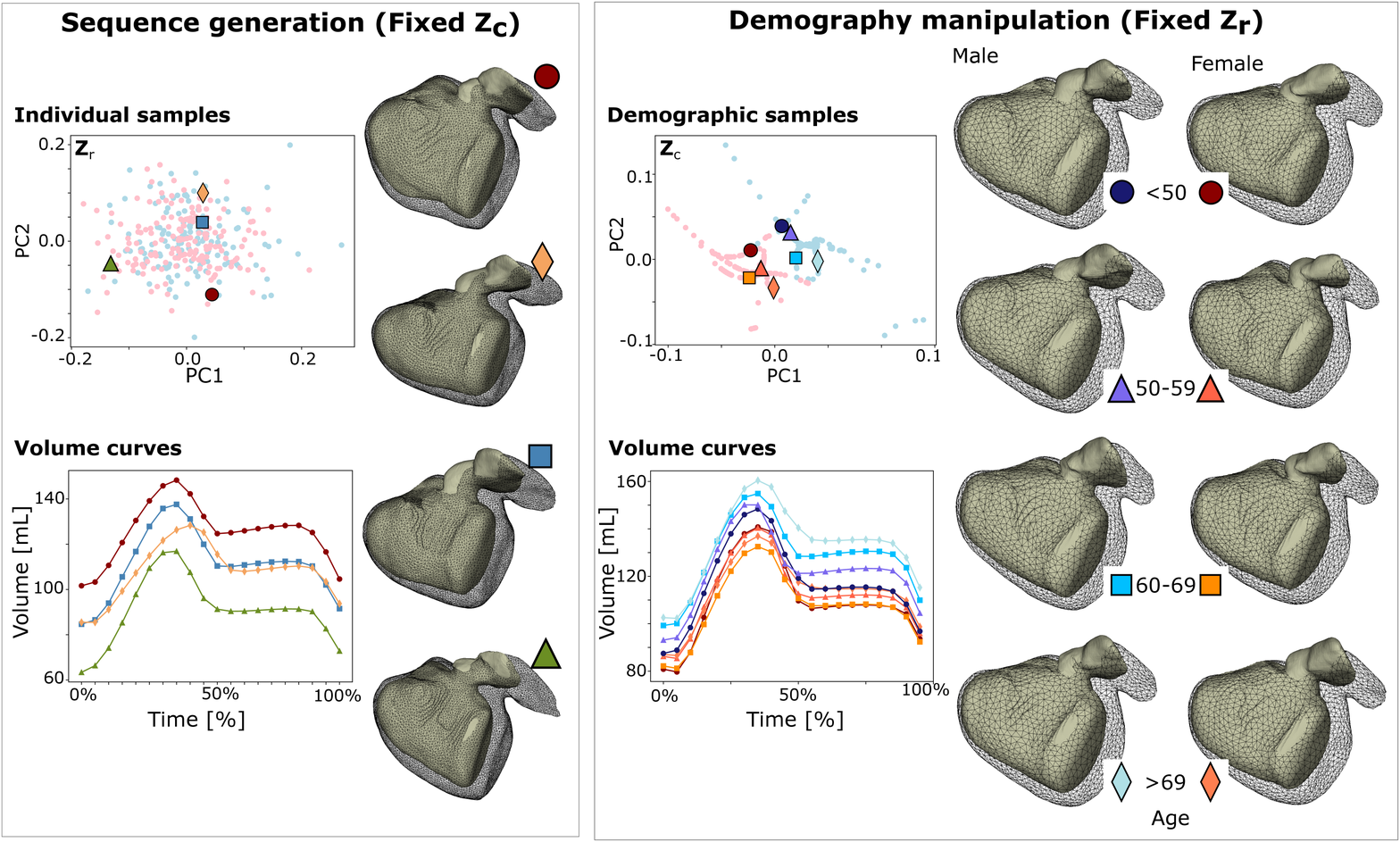}
    \caption{Synthetically created samples. Left: fixed clinical demography (50-59 years old, Male, systolic blood pressure equal to 130 mmHg) and sampled $\textbf{z}_r$. Right: fixed individual latent and varying $\textbf{z}_c$. The figure show the first two principle components (PC) of the latent spaces, the volume curves for all samples as well as the generated anatomies at $t=\%0$ (surface) and at $V_\text{max}$ (wireframe).}
    \label{fig:combined_generation}
\end{figure}

\textbf{Sequence generation ---}
The proposed method can be used to generate unique sequences with specified clinical demography. 
Figure \ref{fig:combined_generation}(left) shows four synthetically created samples based on the same clinical demography. 
We observe that all generated anatomies are continuous and smooth surfaces, with volumes indicating normal atrial motion. 
It is evident that multiple plausible LA and LAA geometries and motions exist despite being generated from the same clinical demography. 
This diversity is attributed to factors not included in the model (ie. height, weight, smoking status, etc.), but also individual traits that cannot be directly related to demography or disease. 
To evaluate the models ability to generate realistic cohorts, we generate a synthetic counterpart to the test set, where a synthetic anatomical sequence is generated based on the clinical demography of each person in the test set. 
Since multiple plausible sequences can be generated based on the same clinical demography, we do not compare the generated surfaces directly to those from the test set. 
Instead, Figure \ref{fig:generated_distributions} shows the distribution of measured FC in both the real and synthetic data in eight subgroups split on age and gender.
Both the real and the generated data exhibit similar trends; the FC decreases with age and are higher in women compared to men.
The longer tails in the real data are attributed outliers errors made by the automatic image segmentation, where the volume is over- or underestimated in one or more time frames. 

%\begin{figure}[tb!]
%    \centering
%    \includegraphics[width=\textwidth]{figs/generated_samples.eps}
%    \caption{Five synthetically created samples representing plausible anatomical sequences from a 50-59 year old male with systolic blood pressure equal to 130 mmHg. We show (A) the first two principal components of the five sampled individual latent vectors together with the training embeddings (male/female in blue/pink), (B) the volume curves and (C) the anatomy at $t=0\%$ (surface) and at max volume (wireframe).}
%    \label{fig:generation}
%\end{figure}

% Visually assure that the generated sequences have an anatomy and motion hereof share a high similarity to the real anatomy and motion. 
% Demonstrate the certain parameters fulfull the same as the real data (volume curves) 
% Confirm some of the known hypotheses between clinical data and shape/motion

\textbf{Demography manipulation ---}
%As a final test for the conditional generative abilities, we investigate the effect of changing the clinical conditions during training. 
We sample a single individual vector $\textbf{z}_r$ and investigate the effect of changing the gender and age in the clinical demography input.
Figure \ref{fig:combined_generation}(right) shows how changing the demography alters the anatomical sequence. 
It can be noted that the shared $\textbf{z}_r$ ensures a relatively stable overall anatomy across all samples, but that associations are captured such that changing the gender from male to female for example results in a smaller LA volume.
The blue/pink opaque circles in the Principle Component Analysis (PCA) plots in Figure \ref{fig:combined_generation} show the embeddings of the training data for male/female participants. 
We observe a clear gender separation in clinical demographic space, whereas the individual latent space shows no obvious gender separation. 
This supports our idea of $\textbf{z}_r$ as a residual latent vector and that the two spaces can be sampled independently. 

%\begin{figure}[tb!]
%    \centering
%    \includegraphics[width=.9\linewidth]{figs/condition_manipulation_v3.eps}
%    \caption{Synthetically created samples with fixed $\textbf{z}_r$ and varying clinical demography. (A) The first two principal components of both latent spaces, (B) the generated anatomies at $t=\%0$ (surface) and at $V_\text{max}$ (wireframe) and (C) the volume curves for each case.}
%    \label{fig:condition_manipulation}
%\end{figure}

\section{Discussion and conclusion}
    We have presented a conditional generative model, that is capable of creating plausible temporal sequences of cardiac anatomy while taking patient demography into account. 
The neural SDF models the cardiac anatomy and movement jointly, and allows for generating smooth and detailed anatomical sequences with close resemblance to real anatomies. 
We showed that the proposed model outperforms a current state-of-the-art method for anatomical sequence completion and that it can be used to estimate functional parameters from the large pool of CT protocols, where only one or two best-phase images are acquired. 
In contrast to previous work, the versatility of the neural SDF representation allows for basing the sequence completion on any single time frame or even multiple scans (i.e. end-systolic and end-diastolic). 
%To address the difficulties in getting the timing right, future work includes adding the heart rate (as measured by the scanner during acquisition) to the clinical demography and consider a sinusoidal time-embedding to account for the repetitive pattern. 
We demonstrated that the model was able to learn abstract associations between the clinical demography and atrial shape and motion. 
We expect that the model can be extended with a larger collection of diverse non-imaging data, which will allow for generating synthetic populations with specific demography, disease status or biophysical constraints. 
Being able to generate realistic anatomical sequences of dynamically moving anatomies based on such conditions is a valuable tool for methods such as fluid simulation, operation planning and disease detection not only in the cardiac domain, but any domain where spatio-temporal models are of interest.

\begin{credits}
\subsubsection{\ackname} This work was supported by a PhD grant from the Technical University of Denmark - Department of Applied Mathematics and Computer Science (DTU Compute) and d by the Spanish Ministry of Science and Innovation under the Programme "Proyectos de Generación de Conocimiento 2022" (PID2022-143239OB-I00).

\subsubsection{\discintname}
Kristine Sørensen now works for Novo Nordisk A/S. Rasmus Paulsen and Klaus Kofoed has received a research grant from Novo Nordisk A/S. 
\end{credits}
%
% ---- Bibliography ----
%
% BibTeX users should specify bibliography style 'splncs04'.
% References will then be sorted and formatted in the correct style.
%
\bibliographystyle{splncs04}
\bibliography{references}
\end{document}